\documentclass[runningheads]{llncs}
\usepackage{graphicx}
\usepackage{comment}
\usepackage{amsmath,amssymb} 
\usepackage{color}

\usepackage{multirow}
\usepackage{bbm}
\usepackage{booktabs}
\usepackage{amssymb}
\usepackage{pifont}

\newcommand{\cmark}{\ding{51}}
\newcommand{\xmark}{\ding{55}}

\usepackage[pagebackref=true,breaklinks=true,letterpaper=true,colorlinks,bookmarks=false]{hyperref}

\def\eg{\emph{e.g}.} 
\def\ie{\emph{i.e}. }

\def\wrt{w.r.t.} 
\def\etal{\emph{et al}.}

\begin{document}
\pagestyle{headings}
\mainmatter
\def\ECCVSubNumber{3183}  

\title{Appearance-Preserving 3D Convolution for Video-based Person Re-identification} 



\titlerunning{Appearance-Preserving 3D Convolution for Video-based Person ReID}

\author{Xinqian Gu\inst{1,2} \and
Hong Chang\inst{1,2} \and
Bingpeng Ma\inst{2} \and \\
Hongkai Zhang\inst{1,2} \and
Xilin Chen\inst{1,2}}
\authorrunning{X. Gu, H. Chang et al.}
%
\institute{	Key Lab of Intelligent Information Processing of Chinese Academy of Sciences (CAS), Institute of Computing Technology, CAS, Beijing, 100190, China \and
University of Chinese Academy of Sciences, Beijing, 100049, China\\
\email{xinqian.gu@vipl.ict.ac.cn, changhong@ict.ac.cn, bpma@ucas.ac.cn, hongkai.zhang@vipl.ict.ac.cn, xlchen@ict.ac.cn}}

\maketitle

\begin{abstract}
Due to the imperfect person detection results and posture changes, temporal appearance misalignment is unavoidable in video-based person re-identification (ReID).
In this case, 3D convolution may destroy the appearance representation of person video clips, thus it is harmful to ReID.
To address this problem, we propose Appearance-Preserving 3D Convolution (AP3D), which is composed of two components: an Appearance-Preserving Module (APM) and a 3D convolution kernel.
With APM aligning the adjacent feature maps in pixel level, the following 3D convolution can model temporal information on the premise of maintaining the appearance representation quality.
It is easy to combine AP3D with existing 3D ConvNets by simply replacing the original 3D convolution kernels with AP3Ds.
Extensive experiments demonstrate the effectiveness of AP3D for video-based ReID and the results on three widely used datasets surpass the state-of-the-arts.
Code is available at: \url{https://github.com/guxinqian/AP3D}.

\end{abstract}

\section{Introduction}

Video-based person re-identification (ReID)~\cite{Wang2014Person,Hou2020TCL,Hou2019vrstc} plays a crucial role in intelligent video surveillance system.
Compared with image-based ReID~\cite{Sun2018Beyond,Hou2019Interaction}, the main difference is that the query and gallery in video-based ReID are both videos and contain
additional temporal information.
Therefore, how to deal with the temporal relations between video frames effectively is of central importance in video-based ReID.

\begin{figure}[t]
	\centering
	\includegraphics[width = 1\columnwidth]{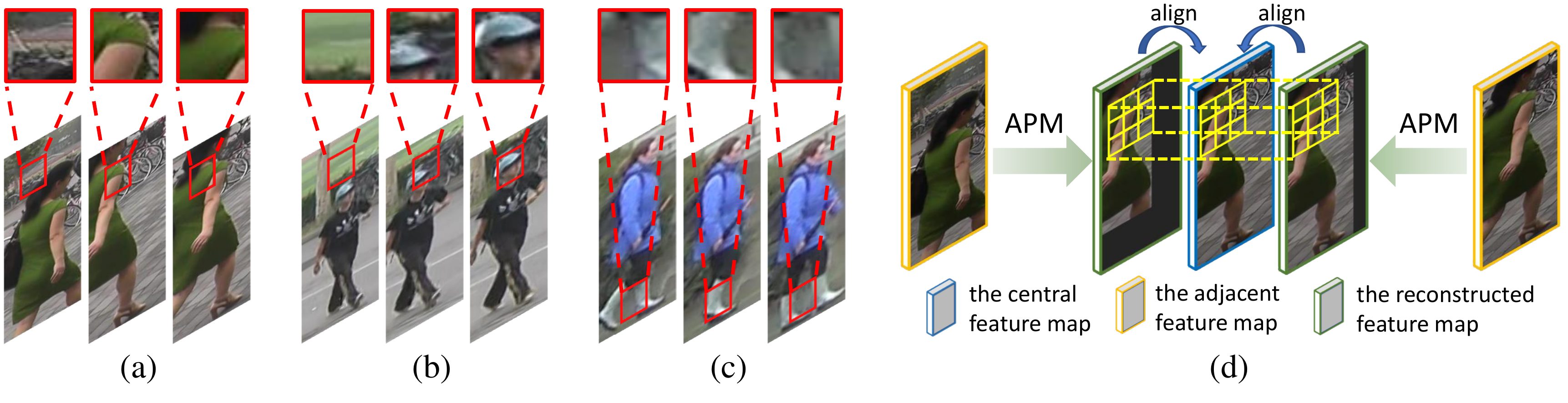}\\
	\caption{Temporal appearance misalignment caused by (a) smaller bounding boxes, (b) bigger bounding boxes and (c) posture changes.
		(d) AP3D firstly uses APM to reconstruct the adjacent feature maps to guarantee the appearance alignment with respect to the central feature map and then performs
		3D convolution}
	\label{fig:introduction}
\end{figure}

The most commonly used temporal information modeling methods in computer vision include LSTM~\cite{Hochreiter1997LSTM,Ng2015Beyond}, 3D convolution~\cite{Tran2015Learning,Carreira2017I3D,Qiu2017P3D}, and Non-local operation~\cite{Wang2018Non}.
LSTM and 3D convolution are adept at dealing with local temporal relations and encoding the relative position.
Some researchers~\cite{Carreira2017I3D} have demonstrated that 3D convolution is superior to CNN+LSTM on the video classification tasks.
In contrast, Non-local operation does not encode the relative position, but it can model long-range temporal dependencies.
These methods are complementary to each other.
In this paper, we mainly focus on improving existing 3D convolution to make it more suitable for video-based ReID.

Recently, some researchers~\cite{Liao2018VideoBasedPR,li2018M3D} try to introduce 3D convolution to video-based ReID.
However, they neglect that, compared with other video-based tasks, the video sample in video-based ReID consists of a sequence of bounding boxes produced by some pedestrian detector~\cite{Ren2015Faster,Zhang2020DRCNN} (see Figure~\ref{fig:introduction}), not the original video frames.
Due to the imperfect person detection algorithm, some resulting bounding boxes are smaller (see Figure~\ref{fig:introduction} (a)) or bigger (see Figure~\ref{fig:introduction} (b)) than the ground truths.
In this case, because of the resizing operation before feeding into a neural network, the same spatial positions in adjacent frames may belong to different body parts and the same body parts in adjacent frames may be scaled to different sizes.
Even though the detection results are accurate, the misalignment problem may still exist due to the posture changes of the target person (see Figure~\ref{fig:introduction} (c)).
Note that one 3D convolution kernel processes the features at the same spatial position in adjacent frames into one value.
When temporal appearance misalignment exists, 3D convolution may mixture the features belonging to different body parts in adjacent frames into one feature, which destroys the appearance representations of person videos.
Since the performance of video-based ReID highly relies on the appearance representation, so the appearance destruction is harmful.
Therefore, it is desirable to develop a new 3D convolution method which can model temporal relations on the premise of maintaining appearance representation quality.

In this paper, we propose Appearance-Preserving 3D convolution (AP3D) to address the appearance destruction problem of existing 3D convolution.
As shown in Figure~\ref{fig:introduction} (d), AP3D is composed of an Appearance-Preserving Module (APM) and a 3D convolution kernel.
For each central feature map, APM reconstructs its adjacent feature maps according to the cross-pixel semantic similarity and guarantees the temporal appearance alignment between the reconstructed and central feature maps.
The reconstruction process of APM can be considered as feature map registration between two frames.
As for the problem of asymmetric appearance information (\eg, in Figure~\ref{fig:introduction} (a), the first frame does not contain foot region, thus can not be aligned
with the second frame perfectly), Contrastive Attention is proposed to find the unmatched regions between the reconstructed and central feature maps.
Then, the learned attention mask is imposed on the reconstructed feature map to avoid error propagation.
With APM guaranteeing the appearance alignment, the following 3D convolution can model the spatiotemporal information more effectively and enhance the video representation with higher discriminative ability but no appearance
destruction.
Consequently, the performance of video-based ReID can be greatly improved.
Note that the learning process of APM is unsupervised. In other words, no extra correspondence annotations are required, and the model can be trained only with identification
supervision.

The proposed AP3D can be easily combined with existing 3D ConvNets (\eg, I3D~\cite{Carreira2017I3D} and P3D~\cite{Qiu2017P3D}) just by replacing the original 3D convolution kernels with AP3Ds.
Extensive ablation studies on two widely used datasets indicate that AP3D outperforms existing 3D convolution significantly .
Using RGB information only and without any bells and whistles (\eg, optical flow, complex feature matching strategy), AP3D achieves state-of-the-art results on both datasets.

In summary, the main contributions of our work lie in three aspects:
(1) finding that existing 3D convolution is problematic for extracting appearance representation when misalignment exists;
(2) proposing an AP3D method to address this problem by aligning the feature maps in pixel level according to semantic similarity before convolution operation;
(3) achieving superior performance on video-based ReID compared with state-of-the-art methods.

\section{Related Work}

\noindent
{\bf Video-based ReID.}
Compared with image-based ReID, the samples in video-based ReID contain more frames and additional temporal information.
Therefore, some existing methods \cite{li2018M3D,Wang2014Person,Chung2017A,Mclaughlin2016Recurrent} attempt to model the additional temporal information to enhance the video
representations.
In contrast, other methods \cite{Liu2017QAN,Li2018Diversity,Si2018Dual,Chen2018Video} extract video frame features just using image-based ReID model and explore how to
integrate or match multi-frame features.
In this paper, we try to solve video-based ReID through developing an improved 3D convolution model for better spatiotemporal feature representation.

\noindent
{\bf Temporal Information Modeling.}
The widely used temporal information modeling methods in computer vision include LSTM \cite{Hochreiter1997LSTM,Ng2015Beyond}, 3D convolution~\cite{Tran2015Learning,Carreira2017I3D}, and Non-local operation \cite{Wang2018Non}.
LSTM and 3D convolution are adept at modeling local temporal relations and encoding the relative position, while Non-local operation can deal with long-range temporal relations.
They are complementary to each other.
Zisserman \etal \cite{Carreira2017I3D} has demonstrated that 3D convolution outperforms CNN+LSTM on the video classification task.
In this paper, we mainly improve the original 3D convolution to avoid the appearance destruction problem and also attempt to combine the proposed AP3D with some existing 3D
ConvNets.

\noindent
{\bf Image Registration.}
Transforming different images into the same coordinate system is called image registration \cite{Barbara2003Image,aberman2018neural}.
These images may be obtained at different times, from different viewpoints or different modalities.
The spatial relations between these images may be estimated using rigid, affine, or complex deformation models.
As for the proposed method, the alignment operation of APM can be considered as feature map registration.
Different feature maps are obtained at sequential times and the subject of person is non-rigid.

\section{Appearance-Preserving 3D Convolution}
In this section, we first illustrate the overall framework of the proposed AP3D.
Then, the details of the core module, \ie Appearance-Preserving Module (APM), are explained followed with discussion.
Finally, we introduce how to combine AP3D with existing 3D ConvNets.

\begin{figure}[t]
	\centering
	\includegraphics[width = 1\linewidth]{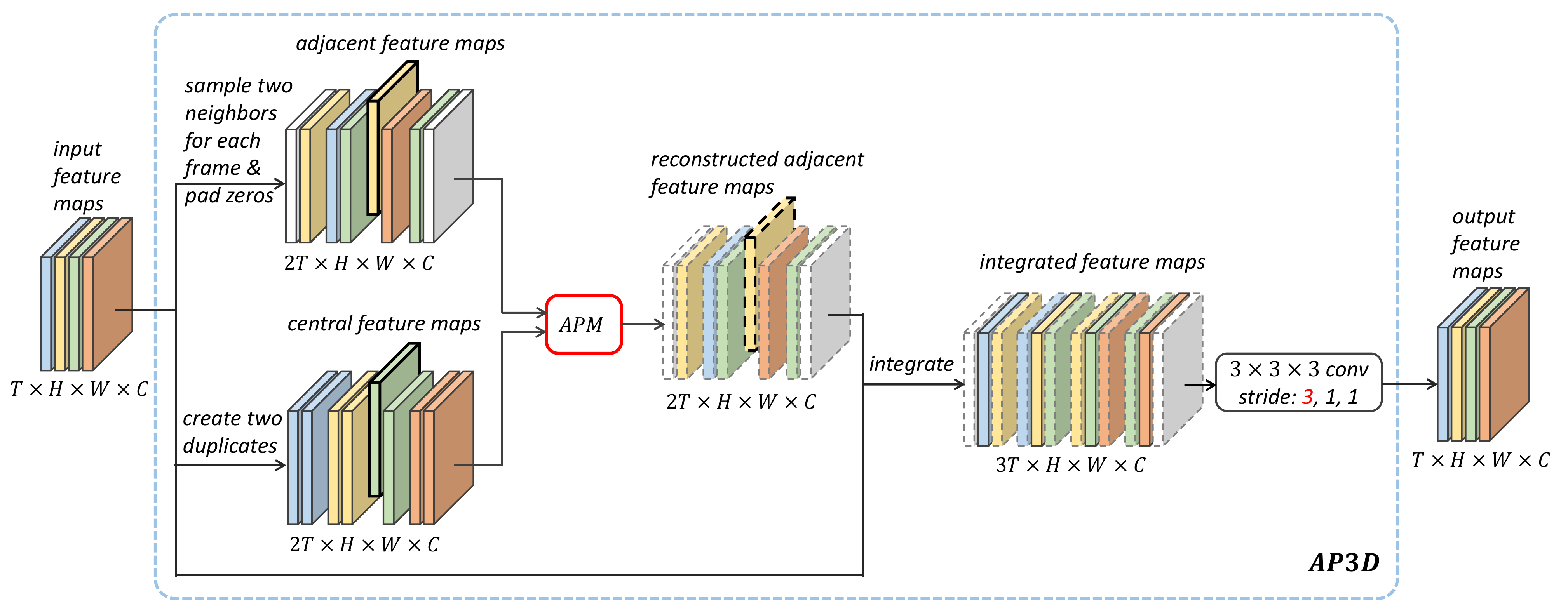}\\
	\caption{The overall framework of the proposed AP3D.
		Each feature map of the input tensor is considered as the central feature map and its two neighbors are sampled as the corresponding adjacent feature maps.
		APM is used to reconstruct the adjacent feature maps to guarantee the appearance alignment with respect to corresponding central feature maps.
		Then the following 3D convolution is performed.
		Note that the temporal stride of 3D convolution kernel is set to its temporal kernel size.
		In that case, the shape of output tensor is the same as the shape of input tensor}
	\label{fig:arch}
\end{figure}

\subsection{The Framework}

3D convolution is widely used on video classification task and achieves state-of-the-art performance.
Recently, some researchers~\cite{Liao2018VideoBasedPR,li2018M3D} introduce it to video-based ReID.
However, they neglect that 
the performance of ReID tasks is highly dependent on the appearance representation, instead of the motion representation.
Due to the imperfect detection results or posture changes, appearance misalignment is unavoidable in video-based ReID samples.
In this case, existing 3D convolutions, which process the same spatial position across adjacent frames as a whole, may destroy the appearance representation of person videos,
therefore they are harmful to ReID.

In this paper, we propose a novel AP3D method to address the above problem.
The proposed AP3D is composed of an APM and a following 3D convolution.
An example of AP3D with $3\times3\times3$ convolution kernel is shown in Figure~\ref{fig:arch}.
Specifically, given an input tensor with $T$ frames, each frame is considered as the central frame. 
We first sample two neighbors for each frame and obtain $2T$ adjacent feature maps in total after padding zeros.
Secondly, APM is used to reconstruct each adjacent feature map to guarantee the appearance alignment with corresponding central feature map.
Then, we integrate the reconstructed adjacent feature maps and the original input feature maps to form a temporary tensor.
Finally, the $3\times3\times3$ convolution with stride (3, 1, 1) is performed and an output tensor with $T$ frames can be produced.
With APM guaranteeing appearance alignment, the following 3D convolution can model temporal relations without  appearance destruction.
The details of APM are presented in next subsection.

\subsection{Appearance-Preserving Module}
\label{sec:APM}

\noindent
{\bf Feature Map Registration.}
The objective of APM is reconstructing each adjacent feature map to guarantee that the same spatial position on the reconstructed and corresponding central feature maps belong to the same body part.
It can be considered as a graph matching or registration task between each two feature maps.
On one hand, since the human body is a non-rigid object, a simple affine transformation can not achieve this goal.
On the other hand, existing video-based ReID datasets do not have extra correspondence annotations.
Therefore, the process of registration is not that straightforward.

\begin{figure}[t]
	\centering
	\includegraphics[width = 0.6\columnwidth]{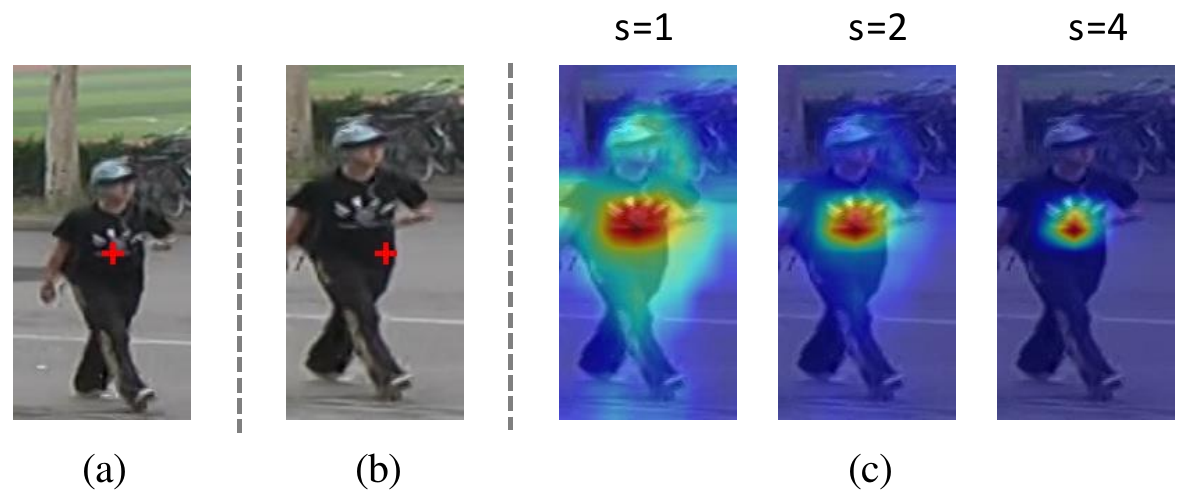}\\
	\caption{Visualization of (a) a central frame, (b) its adjacent frame and (c) similarity distribution with different scale factors $s$ on the adjacent feature maps.
		With a reasonable $s$, APM can locate the corresponding region on the adjacent feature map \wrt the marked position on the central frame accurately}
	\label{fig:visualization}
\end{figure}

We notice that the middle-level features from ConvNet contain some semantic information \cite{aberman2018neural}.
In general, the features with the same appearance have higher cosine similarity, while the features with different appearances have lower cosine similarity \cite{aberman2018neural,Hou2019vrstc}.
As shown in Figure~\ref{fig:visualization}, the red crosses indicate the same position on the central (in Figure~\ref{fig:visualization}~(a)) and adjacent (in Figure~\ref{fig:visualization}~(b)) frames, but they belong to different body parts.
We compute the cross-pixel cosine similarities between the marked position on the central feature map and all positions on the adjacent feature map.
After normalization, the similarity distribution is visualized in Figure~\ref{fig:visualization}~(c) ($s=1$).
It can be seen that the region with the same appearance is highlighted.
Hence, in this paper, we locate the corresponding positions in adjacent frames according to the cross-pixel similarities to achieve feature map registration.

Since the scales of the same body part on the adjacent feature maps may be different, one position on the central feature map may have several corresponding pixels on its adjacent feature map, and vice versa.
Therefore, filling the corresponding position on the reconstructed feature map with only the most similar position on the original adjacent feature map is not accurate.
To include all pixels with the same appearance, we compute the response $y_i$ at each position on the reconstructed adjacent feature map as a weighted sum of the features $x_j$ at all positions on the
original adjacent feature map:
\begin{equation}\label{eq:weightedsum}
y_i=\sum\limits_j \frac{ e^{f(c_i,x_j)}x_j}{\sum\limits_j e^{f(c_i,x_j)}},
\end{equation}
where $c_i$ is the feature on the central feature map with the same spatial position as $y_i$ and $f(c_i,x_j)$ is defined as the cosine similarity between $c_i$ and $x_j$ with a scale factor $s>0$:
\begin{equation}\label{eq:cos}
f(c_i,x_j)=s\ \frac{g(c_i)\cdot g(x_j)}{\lVert g(c_i) \rVert \lVert g(x_j) \rVert},
\end{equation}
where $g(\cdot)$ is a linear transformation that maps the features to a low-dimensional space.
The scale factor $s$ is used to adjust the range of cosine similarities.
And a big $s$ can make the relatively high similarity even higher while the relatively low similarity lower.
As shown in Figure~\ref{fig:visualization}~(c), with a reasonable scale factor $s$, APM can locate the corresponding region on the adjacent feature map precisely.
In this paper, We set the scale factor to 4.

\begin{figure}[t]
	\centering
	\includegraphics[width = \linewidth]{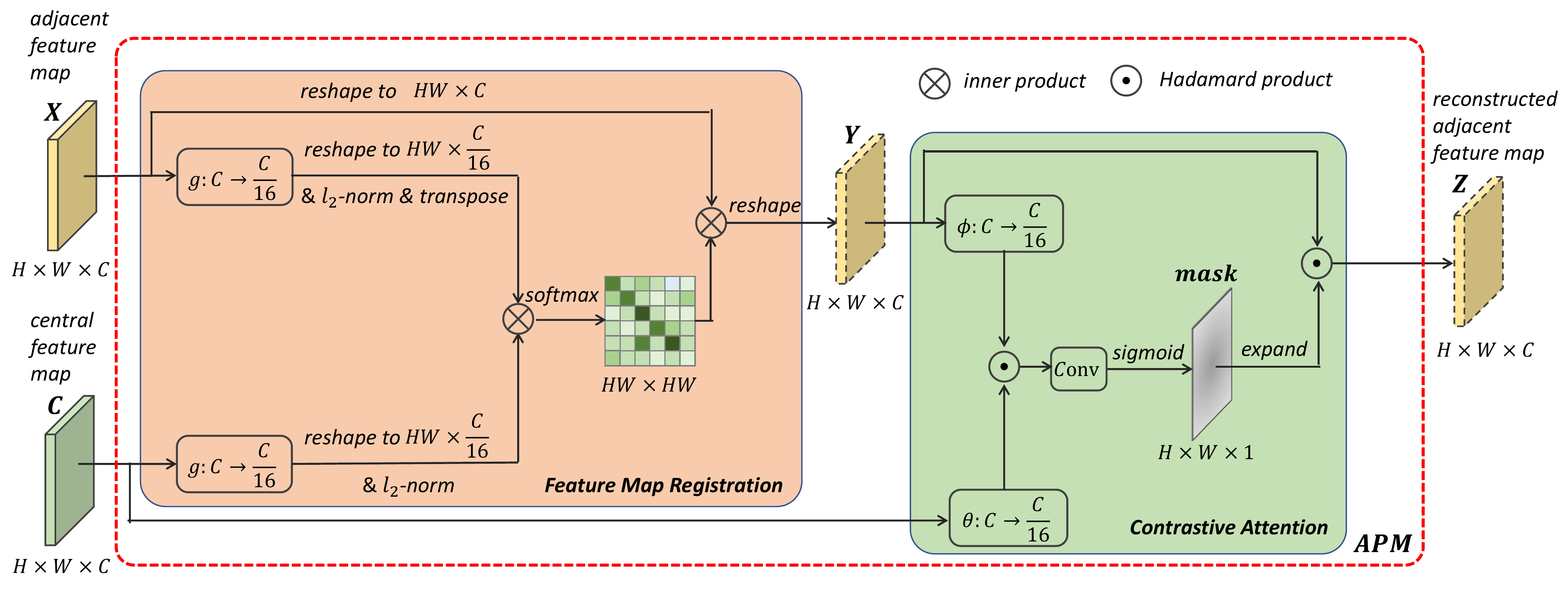}\\
	\caption{The illustration of APM. The adjacent feature map is firstly reconstructed by feature map registration.
		Then a Contrastive Attention mask is multiplied with the reconstructed feature map to avoid error propagation caused by imperfect registration}
	\label{fig:apm}
\end{figure}

\noindent
{\bf Contrastive Attention.}
Due to the error of pedestrian detection, some regressive bounding boxes are smaller than the ground truths, so some body parts may be lost in the adjacent frames (see Figure~\ref{fig:introduction}~(a)).
In this case, the adjacent feature maps can not align with the central feature map perfectly.
To avoid error propagation caused by imperfect registration, Contrastive Attention is proposed to find the unmatched regions between the reconstructed and central feature maps.
Then, the learned attention mask is imposed on the reconstructed feature map.
The final response $z_i$ at each position on the reconstructed feature map is defined as:
\begin{equation}\label{eq:mask}
z_i=ContrastiveAtt(c_i, y_i) y_i.
\end{equation}
Here $ContrastiveAtt(c_i, y_i)$ produces an attention value in $[0,1]$ accoring to the semantic similarity between $c_i$ and $y_i$:
\begin{equation}\label{eq:maskg}
ContrastiveAtt(c_i, y_i)=sigmoid(w^T(\theta(c_i)\odot \phi(y_i))),
\end{equation}
where $w$ is a learnable weight vector implemented by $1\times1$ convolution, and $\odot$ is Hadamard product.
Since $c_i$ and $y_i$ are from the central and reconstructed feature maps respectively, we use two asymmetric mapping functions $\theta(\cdot)$ and $\phi(\cdot)$ to map $c_i$
and $y_i$ to a shared low-dimension semantic space.

The registration and contrastive attention of APM are illustrated in Figure~\ref{fig:apm}.
All three semantic mappings, \ie $g$, $\theta$ and $\phi$, are implemented by $1\times1$ convolution layers.
To reduce the computation, the output channels of these convolution layers are set to $C/16$.

\subsection{Discussion}

\noindent
{\bf Relations between APM and Non-local.}
APM and Non-local (NL) operation can be viewed as two graph neural network modules.
Both modules consider the feature at each position on feature maps as a node in graph and use weighted sum to estimate the feature.
But they have many differences:
	
(a) NL aims to use spatiotemporal information to enhance feature and its essence is graph convolution or self-attention on a spatiotemporal graph.
In contrast, APM aims to reconstruct adjacent feature maps to avoid appearance destruction by the following 3D Conv.
Its essence is graph matching or registration between two spatial graphs.
	
(b) The weights in the weighted sum in NL are used for building dependencies between each pair of nodes only and do not have specific meaning.
In contrast, APM defines the weights using cosine similarity with a reasonable scale factor, in order to find the positions with the same appearance on the adjacent feature maps accurately (see Figure~\ref{fig:visualization}).
	
(c) After APM, the integrated feature maps in Figure~\ref{fig:arch} can still maintain spatiotemporal relative relations to be encoded by the following 3D Conv, while NL cannot.
	
(d) Given a spatiotemporal graph with $N$ frames, the computational complexity of NL is $O(N^2)$, while the computational complexity of APM is only $O(N)$, much lower than NL.

\noindent
{\bf Relations between Contrastive Attention and Spatial Attention.}
The Contrastive Attention in APM aims to find the unmatched regions between two frames to avoid error propagation caused by imperfect registration, while the widely used spatial attention~\cite{Li2018Diversity} in ReID aims to locate more discriminative regions for each frame.
As for formulation, Contrastive Attention takes two feature maps as inputs and is imposed on the reconstructed feature map, while Spatial Attention takes one feature map as input and is imposed on itself.

\begin{figure}[t]
	\centering
	\includegraphics[width = 1\columnwidth]{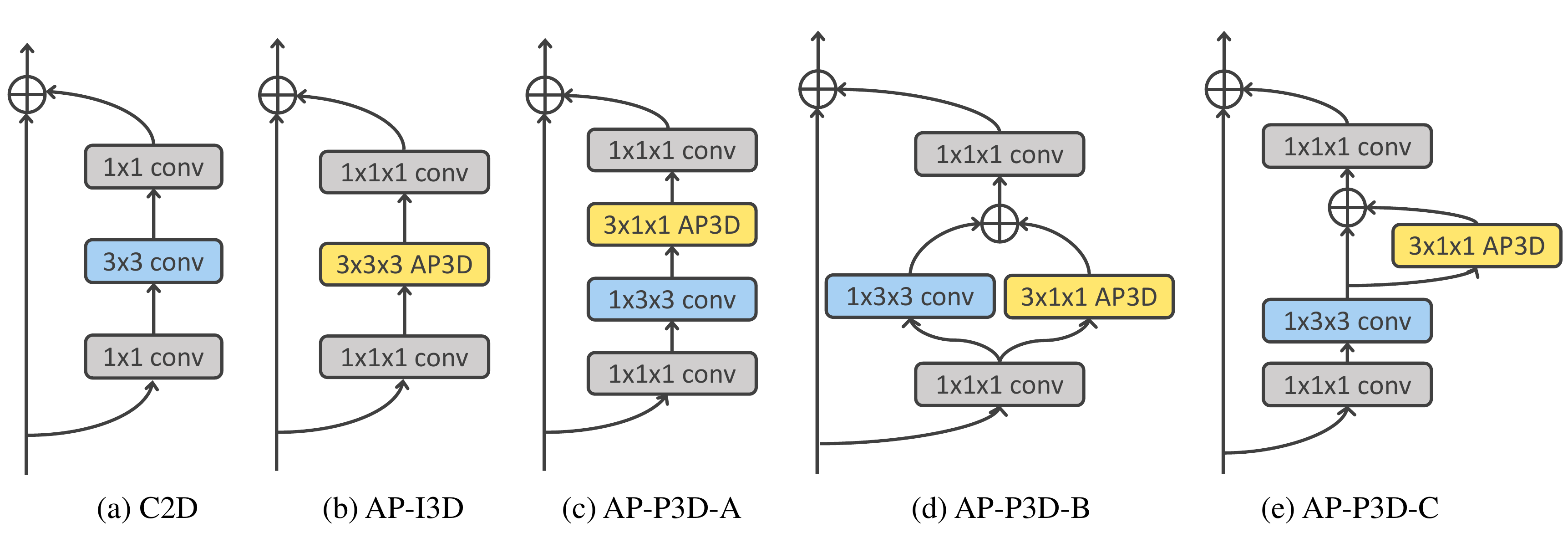}\\
	\caption{The C2D, AP-I3D and AP-P3D versions of Residual blocks.
		As for AP-I3D and AP-P3D Residual blocks, only the origional temporal convolution kernels are replaced by AP3Ds}
	\label{fig:AP3D}
\end{figure}

\subsection{Combining AP3D with I3D and P3D Blocks}

To leverage successful 3D ConvNet designs, we combine the proposed AP3D with I3D~\cite{Carreira2017I3D} and P3D~\cite{Qiu2017P3D} Residual blocks.
Transferring I3D and P3D Residual blocks to their AP3D versions just needs to replace the original temporal convolution kernel with AP3D with the same kernel size.
The C2D, AP-I3D and AP-P3D versions of Residual blocks are shown in Figure~\ref{fig:AP3D}.

\section{AP3D for Video-based ReID}
To investigate the effectiveness of AP3D for video-based ReID, we use the 2D ConvNet (C2D) form \cite{Hou2019vrstc} as our baseline method and extend it into AP3D ConvNet with the proposed AP3D.
The details of network architectures are described in Section \ref{sec:network}, and then the loss function we use is introduced in Section \ref{sec:loss}.

\subsection{Network Architectures}
\label{sec:network}

\noindent
{\bf C2D baseline.}
We use ResNet-50~\cite{He2016Deep} pre-trained on ImageNet~\cite{Olga2015ImageNet} as the backbone and remove the down-sampling operation of $\text{stage}_5$ following
\cite{Sun2018Beyond} to enrich the granularity.
Given an input video clip with $T$ frames, it outputs a tensor with shape $T\times H \times W \times 2048$.
After spatial max pooling and temporal average pooling, a 2048-dimension feature is produced.
Before feeding into the classifier, a BatchNorm~\cite{Ioffe2015BN} operation is used to normalize the feature following \cite{Hou2019vrstc}.
The C2D baseline does not involve any temporal operations except the final temporal average pooling.

\noindent
{\bf AP3D ConvNet.}
We replace some 2D Residual blocks with AP3D Residual blocks to turn C2D into AP3D ConvNet for spatiotemporal feature learning.
Specifically, we investigate replacing one, half of or all Residual blocks in one stage of ResNet, and the results are reported in Section~\ref{sec:ablation_study}

\subsection{Objective Function}
\label{sec:loss}
Following \cite{Wang2018MGN}, we combine cross entropy loss and triplet loss~\cite{Hermans2017In} for spatiotemporal representation learning. 
Since cross entropy loss mainly optimizes the features in angular subspace~\cite{Wang2018cosface}, to maintain consistency, we use cosine distance for triplet loss.

\section{Experiments}
\subsection{Datasets and Evaluation Protocol}
\noindent
{\bf Datasets.}
We evaluate the proposed method on three video-based ReID datasets, \ie MARS~\cite{Zheng2016MARS}, DukeMTMC-VideoReID~\cite{Wu2018One} and iLIDS-VID~\cite{Wang2014Person}.
Since MARS and DukeMTMC-VideoReID have fixed train/test splits, for convenience, we perform ablation studies mainly on these two datasets.
Besides, we report the final results on iLIDS-VID to compare with the state-of-the-arts.

\noindent
{\bf Evaluation Protocol.}
We use the Cumulative Matching Characteristics (CMC) and mean Average Precision (mAP)~\cite{Zheng2015Scalable} as the evaluation metrics.

\subsection{Implementation Details}
\noindent
{\bf Training.}
In the training stage, for each video tracklet, we randomly sample 4 frames with a stride of 8 frames to form a video clip.
Each batch contains 8 persons, each person with 4 video clips.
We resize all the video frames to $256\times 128$ pixels and use horizontal flip for data augmentation.
As for the optimizer, Adam~\cite{Kingma2014Adam} with weight decay 0.0005 is adopted to update the parameters.
We train the model for 240 epochs in total.
The learning rate is initialized to $3\times 10^{-4}$ and multiplied by 0.1 after every 60 epochs.

\noindent
{\bf Testing.}
In the test phase, for each video tracklet, we first split it into several 32-frame video clips.
Then we extract the feature representation for each video clip and the final video feature is the averaged representation of all clips.
After feature extraction, the cosine distances between the query and gallery features are computed, based on which the retrieval is performed.

\subsection{Comparison with Related Approaches}

\begin{table}[t]
	\centering
	\caption{Comparison between AP3D and original 3D convolution}
	\small
	\begin{center}
		\setlength{\tabcolsep}{1.2mm}{
			\begin{tabular}{l c c c c c c}
				\toprule
				\multirow{2}*{Model} &\multirow{2}*{Param.} &\multirow{2}*{GFLOPs} &\multicolumn{2}{c}{MARS}  &\multicolumn{2}{c}{Duke-Video} \\
				&& &top-1  &mAP &top-1  &mAP\\
				\midrule
				C2D      &23.51 &16.35 &88.9 &83.4 &95.6 &95.1\\
				\midrule
				I3D      &27.64 &19.37 &88.6 &83.0 &95.4 &95.2 \\
				AP-I3D   &27.68 &19.48 &\bfseries90.1 &84.8 &96.2 &95.4 \\
				\midrule
				P3D-A      &24.20 &16.85 &88.9 &83.2 &95.0 &95.0 \\
				AP-P3D-A   &24.24 &16.90 &\bfseries90.1 &84.9 &96.0 &95.3 \\
				\midrule
				P3D-B      &24.20 &16.85 &88.8 &83.0 &95.4 &95.3 \\
				AP-P3D-B   &24.24 &16.96 &89.9 &84.7 &\bfseries96.4 &\bfseries95.9 \\
				\midrule
				P3D-C      &24.20 &16.85 &88.5 &83.1 &95.3 &95.3 \\
				AP-P3D-C   &24.24 &16.90 &\bfseries90.1 &\bfseries85.1 &96.3 &95.6 \\
				\bottomrule
		\end{tabular}}
	\end{center}
	\label{tab:ap3d_vs_3d}
\end{table}

\noindent
{\bf AP3D vs. original 3D convolution.}
To verify the effectiveness and generalization ability of the proposed AP3D, we implement I3D and P3D residual blocks using AP3D and the original 3D convolution,
respectively.
Then, we replace one 2D block with 3D block for every 2 residual blocks in $\text{stage}_2$ and $\text{stage}_3$ of C2D ConvNets, and 5 residual blocks in total are replaced.
As shown in Table~\ref{tab:ap3d_vs_3d}, compared with the C2D baseline, I3D and P3D show close or lower results due to appearance destruction.
With APM aligning the appearance representation, the corresponding AP3D versions improve the performance significantly and consistently on both two datasets with few
additional parameters and little extra computational complexity.
Specifically, AP3D increases about 1\% top-1 and 2\% mAP over I3D and P3D on MARS dataset.
Note that the mAP improvement on DukeMTMC-VideoReID is not as much as that on MARS. One possible explanation is that the bounding boxes of video samples in DukeMTMC-VideoReID
dataset are manually annotated and the appearance misalignment is not too serious, so the improvement of AP3D is not very significant.

Compared with other varieties, AP-P3D-C achieves the best performance among most settings. So we conduct the following experiments based on AP-P3D-C (denoted as AP3D for short) if not specifically noted.

\begin{table}[t]
	\centering
	\caption{Comparison with NL and other temporal information modeling methods}
	\small
	\begin{center}
		\setlength{\tabcolsep}{1.2mm}{
			\begin{tabular}{l c c c c c c}
				\toprule
				\multirow{2}*{Model} &\multirow{2}*{Param.} &\multirow{2}*{GFLOPs} &\multicolumn{2}{c}{MARS}  &\multicolumn{2}{c}{Duke-Video} \\
				&& &top-1  &mAP &top-1  &mAP\\
				\midrule
				NL   &30.87 &21.74    &89.6 &85.0 &96.2 &95.6 \\
				CA-NL &32.75 &21.92     &89.6 &85.0 &95.9 &95.6\\
				NL-P3D  &31.56 &22.17     &89.9 &84.8 &96.2 &95.5\\				
				\midrule
				AP3D &24.24 &16.90   &90.1 &85.1 &96.3 &95.6 \\
				NL-AP3D  &31.60 &22.29     &\bfseries90.7 &\bfseries85.6 &\bfseries97.2 &\bfseries96.1\\
				\midrule
				Deformable 3D Conv~\cite{Dai2017Deformable}    &27.75&19.53        &88.5 &81.9 &95.2 &95.0 \\
				CNN+LSTM~\cite{Hochreiter1997LSTM}   &28.76&16.30  &88.7 &79.8 &95.7 &94.6\\	
				\bottomrule
		\end{tabular}}
	\end{center}
	\label{tab:ap3d_vs_nl}
\end{table}

\noindent
{\bf AP3D vs. Non-local.}
Both APM in AP3D and Non-local (NL) are graph-based methods.
We insert the same 5 NL blocks into C2D ConvNets and compare AP3D with NL in Table~\ref{tab:ap3d_vs_nl}.
It can be seen that, with fewer parameters and less computational complexity, AP3D outperforms NL on both two datasets.

To compare more fairly, we also implement Contrastive Attention embedded Non-local (CA-NL) and the combination of NL and P3D (NL-P3D).
As shown in Table~\ref{tab:ap3d_vs_nl}, CA-NL achieves the same result as NL on MARS and is still inferior to AP3D.
On DukeMTMC-VideoReID, the top-1 of CA-NL is even lower than NL.
It is more likely that the Contrastive Attention in APM is designed to avoid error propagation caused by imperfect registration.
However, the essence of NL is graph convolution on a spatiotemporal graph, not graph registration.
So NL can not co-work with Contrastive Attention.
Besides, since P3D can not handle appearance misalignment in video-based ReID, NL-P3D shows close results to
NL and is inferior to AP3D, too.
With APM aligning the appearance, further improvement is achieved by NL-AP3D.
This result demonstrates that AP3D and NL are complementary to each other.

\noindent
{\bf AP3D vs. other methods for temporal information modeling.}
We also compare AP3D with Deformable 3D convolution~\cite{Dai2017Deformable} and CNN+LSTM~\cite{Hochreiter1997LSTM}.
To compare fairly, the same backbone and hyper-parameters are used.
As shown in Table~\ref{tab:ap3d_vs_nl}, AP3D outperforms these two methods significantly on both two datasets.
This comparison further demonstrates the effectiveness of AP3D for learning temporal cues.

\subsection{Ablation Study}
\label{sec:ablation_study}

\begin{table}[t]
	\centering
	\caption{The results of replacing different numbers of residual blocks in different stages with AP3D block}
	\small
	\begin{center}
		\setlength{\tabcolsep}{2mm}{
			\begin{tabular}{l l c c  c c c}
				\toprule
				\multirow{2}*{Model} &\multirow{2}*{Stage} &\multirow{2}*{Num.} &\multicolumn{2}{c}{MARS}  &\multicolumn{2}{c}{Duke-Video} \\
				& & &top-1  &mAP &top-1  &mAP\\
				\midrule
				C2D   &  &  &88.9 &83.4 &95.6 &95.1\\
				\midrule
				P3D   &$\text{stage}_{2,3}$  &5  &88.5 &83.1 &95.3 &95.3\\
				\midrule
				\multirow{7}*{AP3D}
				&$\text{stage}_1$      &1  &89.0 &83.2 &95.3 &95.1 \\
				&$\text{stage}_2$      &1  &89.5 &84.0 &95.6 &\bfseries95.4 \\
				&$\text{stage}_3$      &1  &\bfseries89.7 &\bfseries84.1 &\bfseries95.9 &95.3 \\
				&$\text{stage}_4$      &1  &88.8 &82.9 &95.4 &95.0 \\
				\cline{2-7}
				&$\text{stage}_{2,3}$  &2  &\bfseries90.1 &84.7 &96.2 &95.4 \\
				&$\text{stage}_{2,3}$  &5  &\bfseries90.1 &\bfseries85.1 &\bfseries96.3 &\bfseries95.6 \\
				&$\text{stage}_{2,3}$  &10 &89.8 &84.7 &95.9 &95.2 \\
				\bottomrule
		\end{tabular}}
	\end{center}
	\label{tab:num}
\end{table}

\noindent
{\bf Effective positions to place AP3D blocks.}
Table~\ref{tab:num} compares the results of replacing a residual block with AP3D block in different stages of C2D ConvNet.
In each of these stages, the second last residual block is replaced with the AP3D block.
It can be seen that the improvements by placing AP3D block in $\text{stage}_2$ and $\text{stage}_3$ are similar.
Especially, the results of placing only one AP3D block in $\text{stage}_2$ or $\text{stage}_3$ surpass the results of placing 5 P3D blocks in $\text{stage}_{2,3}$.
However, the results of placing AP3D block in $\text{stage}_1$ or $\text{stage}_4$ are worse than the C2D baseline.
It is likely that the low-level features in $\text{stage}_1$ are insufficient to provide precise semantic information, thus APM in AP3D can not align the appearance
representation very well.
In contrast, the features in $\text{stage}_4$ are insufficient to provide precise spatial information, so the improvement by appearance alignment is also limited.
Hence, we only consider replacing the residual blocks in $\text{stage}_2$ and $\text{stage}_3$.

\begin{table} [t]
	\begin{minipage}[t]{0.55\linewidth} 
		\centering
		\caption{The results with different backbones}
		\small
		\begin{center}
			\setlength{\tabcolsep}{1mm}{
				\begin{tabular}{l c c c c c}
					\toprule
					\multirow{2}*{Backbone} &\multirow{2}*{Model} &\multicolumn{2}{c}{MARS}  &\multicolumn{2}{c}{Duke-Video} \\
					& &top-1  &mAP &top-1  &mAP\\
					\midrule
					\multirow{3}*{ResNet-18} &C2D    &86.9 &79.0 &93.7 &92.9 \\
					&P3D    &86.9 &79.5 &93.2 &92.9 \\
					&AP3D   &\bfseries88.1 &\bfseries80.9 &\bfseries94.2 &\bfseries93.4 \\
					\midrule
					\multirow{3}*{ResNet-34} &C2D    &87.5 &80.9 &94.6 &93.6 \\
					&P3D    &87.6 &81.0 &94.4 &93.7 \\
					&AP3D   &\bfseries88.7 &\bfseries82.1 &\bfseries95.2 &\bfseries94.7 \\
					\bottomrule
			\end{tabular}}
		\end{center}
		\label{tab:backbone}
	\end{minipage}
	\begin{minipage}[t]{0.45\linewidth} 
		\centering
		\caption{The results of AP3D with/without CA on MARS}
		\small
		\begin{center}
			\setlength{\tabcolsep}{1mm}{
				\begin{tabular}{l c c c}
					\toprule
					Model &w/ CA? &top-1  &mAP\\
					\midrule
					I3D      				&-  &88.6 &83.0  \\
					\multirow{2}*{AP-I3D}   &\xmark  &89.7 &84.7 \\
					&\cmark  &\bfseries90.1 &\bfseries84.8 \\
					\midrule
					P3D      				&- &88.5 &83.1 \\
					\multirow{2}*{AP-P3D} 	&\xmark &89.6 &84.8 \\
					&\cmark &\bfseries90.1 &\bfseries85.1 \\
					\bottomrule
			\end{tabular}}
		\end{center}
		\label{tab:mask}
	\end{minipage} 
\end{table}

\noindent
{\bf How many blocks should be replaced by AP3D?}
Table~\ref{tab:num} also shows the results with more AP3D blocks.
We investigate replacing 2 blocks (1 for each stage), 5 blocks (half of residual blocks in $\text{stage}_2$ and $\text{stage}_3$) and 10 blocks (all residual blocks in
$\text{stage}_2$ and $\text{stage}_3$) in C2D ConvNet.
It can be seen that more AP3D blocks generally lead to higher performance.
We argue that more AP3D blocks can perform more temporal communications, which can hardly be realized via the C2D model.
As for the results with 10 blocks, the performance drop may lie in the overfitting caused by the excessive parameters.

\noindent
{\bf Effectiveness of AP3D across different backbones.}
We also investigate the effectiveness and generalization ability of AP3D across different backbones.
Specifically, we replace half of the residual blocks in $\text{stage}_{2,3}$ of ResNet-18 and ResNet-34 with AP3D blocks.
As shown in Table~\ref{tab:backbone}, AP3D can improve the results of these two architectures significantly and consistently on both datasets.
In particular, AP3D-ResNet-18 is superior to both its ResNet-18 counterparts (C2D and P3D) and the deeper ResNet-34, a model which has almost double the number of parameters and computational complexity, on MARS dataset.
This comparison shows that the effectiveness of AP3D does not rely on additional parameters and computational load.

\noindent
{\bf The effectiveness of Contrastive Attention.}
As described in Section~\ref{sec:APM}, we use Contrastive Attention to avoid error propagation of imperfect registration caused by asymmetric appearance information.
To verify the effectiveness, we reproduce AP3D with/without Contrastive Attention (CA) and the experimental results on MARS, a dataset produced by pedestrian detector, are shown in Table~\ref{tab:mask}.
It can be seen that, without Contrastive Attention, AP-I3D and AP-P3D can still increase the performance of I3D and P3D baselines by a considerable margin. With Contrastive Attention applied on the reconstructed feature map, the results of AP-I3D and AP-P3D can be further improved.

\begin{figure} [t]
	\begin{minipage}[t]{0.46\linewidth} 
		\centering
		\includegraphics[width = 1\columnwidth]{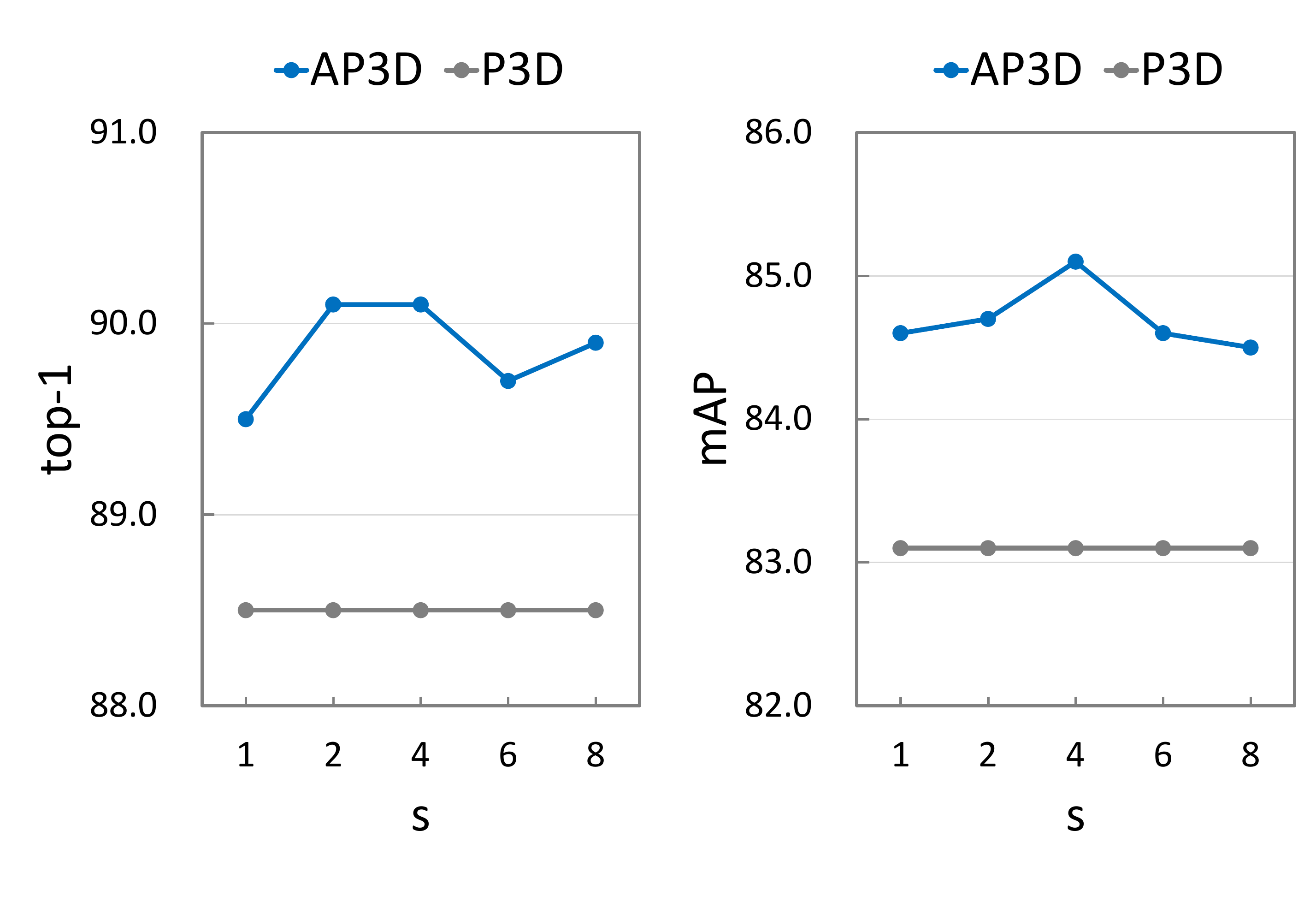}\\
		\caption{The results with different $s$ on MARS dataset}
		\label{fig:scale}
	\end{minipage}
    ~
	\begin{minipage}[t]{0.53\linewidth} 
		\centering
		\includegraphics[width = 1\columnwidth]{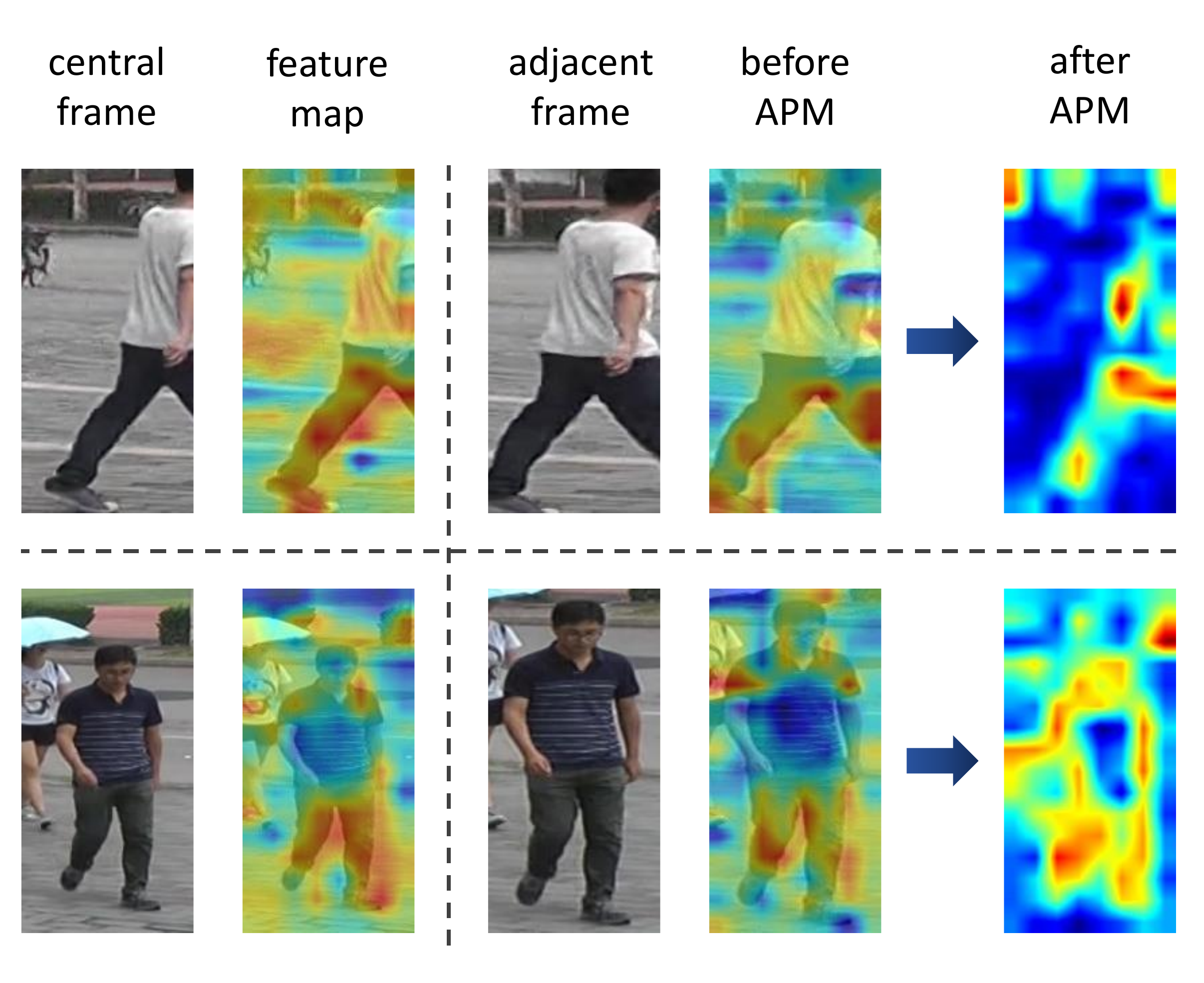}\\
		\caption{The visualization of the original and the reconstructed feature maps after APM}
		\label{fig:vis_case}
	\end{minipage} 
\end{figure}

\noindent
{\bf The influence of the scale factor $s$.}
As discussed in Section~\ref{sec:APM}, the larger the scale factor $s$, the higher the weights of pixels with high similarity.
We show the experimental results with varying $s$ on MARS dataset in Figure~\ref{fig:scale}.
It can be seen that AP3D with different scale factors consistently improves over the baseline and the best performance is achieved when $s=4$.

\begin{table}[t]
	\centering
	\caption{Comparison with state-of-the-arts on MARS, DukeMTMC-VideoReID and iLIDS-VID datasets.
		`Flow' denotes optical flow and `Att.' represents attribute}
	\begin{center}
		\setlength{\tabcolsep}{2mm}{
			\begin{tabular}{l l c c c c c}
				\toprule
				\multirow{2}*{Method}  &\multirow{2}*{Modality} &\multicolumn{2}{c}{MARS}  &\multicolumn{2}{c}{Duke-Video}  &iLIDS-VID \\
				& &top-1  &mAP &top-1  &mAP &top-1 \\
				\midrule
				EUG~\cite{Wu2018One}           &RGB &80.8 &67.4 &83.6 &78.3 &\\
				DuATM~\cite{Si2018Dual}        &RGB &81.2 &67.7 &- &- &-\\
				DRSA~\cite{Li2018Diversity}    &RGB &82.3 &65.8 &-    &-    &80.2\\
				TKP~\cite{gu2019TKP}           &RGB &84.0 &73.3 &94.0 &91.7 &-\\
				M3D~\cite{li2018M3D}           &RGB &84.4 &74.1 &-    &-    &74.0\\
				Snippet~\cite{Chen2018Video}   &RGB + Flow &86.3 &76.1 &- &- &85.4\\
				STA~\cite{Fu2019STA}           &RGB &86.3 &80.8 &96.2 &94.9 &-\\
				AttDriven~\cite{Zhao2019Attribute} &RGB + Att. &87.0 &78.2 &- &- &86.3\\
				GLTR~\cite{Li2019GLTR}         &RGB &87.0 &78.5 &\bfseries96.3 &93.7 &86.0\\
				VRSTC~\cite{Hou2019vrstc}      &RGB &88.5 &82.3 &95.0 &93.5 &83.4\\
				NVAN~\cite{liu2019spatially}   &RGB &90.0 &82.8 &\bfseries96.3 &94.9 &-\\
				\midrule
				AP3D                           &RGB &\bfseries90.1 &\bfseries85.1 &\bfseries96.3 &\bfseries95.6 &\bfseries86.7\\
				NL-AP3D                        &RGB &\bfseries90.7 &\bfseries85.6 &\bfseries97.2 &\bfseries96.1 &\bfseries88.7\\
				\bottomrule
		\end{tabular}}
	\end{center}
	\label{tab:sota}
\end{table}

\subsection{Visualization}
We select some misaligned samples and visualize the original feature maps and the reconstructed feature maps in $\text{stage}_3$ after APM in Figure~\ref{fig:vis_case}.
It can be seen that the highlighted regions of the central feature map and the adjacent feature map before APM mainly focus on their own foreground respectively and are misaligned.
After APM, the highlighted regions of the reconstructed feature maps are aligned \wrt the foreground of the corresponding central frame.
It can further validate the alignment mechanism of APM. 

\subsection{Comparison with State-of-the-Art Methods}

We compare the proposed method with state-of-the-art video-based ReID methods which use the same backbone on MARS, DukeMTMC-VideoReID, and iLIDS-VID datasets.
The results are summarized in Table~\ref{tab:sota}.
Note that these comparison methods differ in many aspects, \eg, using information from different modalities.
Nevertheless, using RGB only and with a simple feature integration strategy (\ie temporal average pooling), the proposed AP3D surpasses all these methods consistently on these three datasets.
Especially, AP3D achieves 85.1\% mAP on MARS dataset.
When combined with Non-local, further improvement can be obtained.

\section{Conclusion}
In this paper, we propose a novel AP3D method for video-based ReID.
AP3D consists of an APM and a 3D convolution kernel.
With APM guaranteeing the appearance alignment across adjacent feature maps, the following 3D convolution can model temporal information on the premise of maintaining the
appearance representation quality.
In this way, the proposed AP3D addresses the appearance destruction problem of the original 3D convolution. It is easy to combine AP3D with existing 3D ConvNets.
Extensive experiments verify the effectiveness and generalization ability of AP3D, which surpasses start-of-the-art methods on three widely used datasets.
As a future work, we will extend AP3D to make it a basic operation in deep neural networks for various video-based recognition tasks.

~

{\bf Acknowledgement} This work is partially supported by Natural Science Foundation of China (NSFC):  61876171 and 61976203.


\clearpage
%
%
\bibliographystyle{splncs04}
\bibliography{egbib}
\end{document}